\documentclass[letterpaper, 10 pt, conference]{ieeeconf}  

\IEEEoverridecommandlockouts                              

\usepackage{amssymb}
\usepackage{amsmath}
\usepackage{booktabs}
\usepackage{graphicx}
\usepackage{multirow}
\usepackage{bigdelim}
\usepackage{pifont}
\usepackage{bbding}
\usepackage{titlesec}
\usepackage{tabularx}
\usepackage{cuted}
\usepackage{caption}
\title{\LARGE \bf
Learn Weightlessness: Imitate Non-Self-Stabilizing Motions on Humanoid Robot
}

\author{%
}
\author{Yucheng Xin$^{1,2*}$, Jiacheng Bao$^{2*}$, Haoran Yang$^{3,2*}$, Wenqiang Que$^{2}$, Dong Wang$^{2\dagger}$\\
Junbo Tan$^{1}$, Xueqian Wang$^{1}$, Bin Zhao$^{2}$, Xuelong Li$^{2}$
\thanks{$^{1}$Center for Artificial Intelligence and Robotics, Shenzhen International Graduate School, Tsinghua University, China, {\tt\small \{xin-yc23@mails., wang.xq@sz., tjblql@sz.\}tsinghua.edu.cn}, $^{2}$Shanghai AI Laboratory, $^{3}$University of Science and Technology of China}
\thanks{$^\dagger$Corresponding author, $^{*}$indicates equal contribution}
}

\begin{document}

\maketitle
\thispagestyle{empty}
\pagestyle{empty}

\begin{strip}
\vspace{-1.5cm}
\begin{center}
    \centering
      \includegraphics[width=1.0\linewidth]{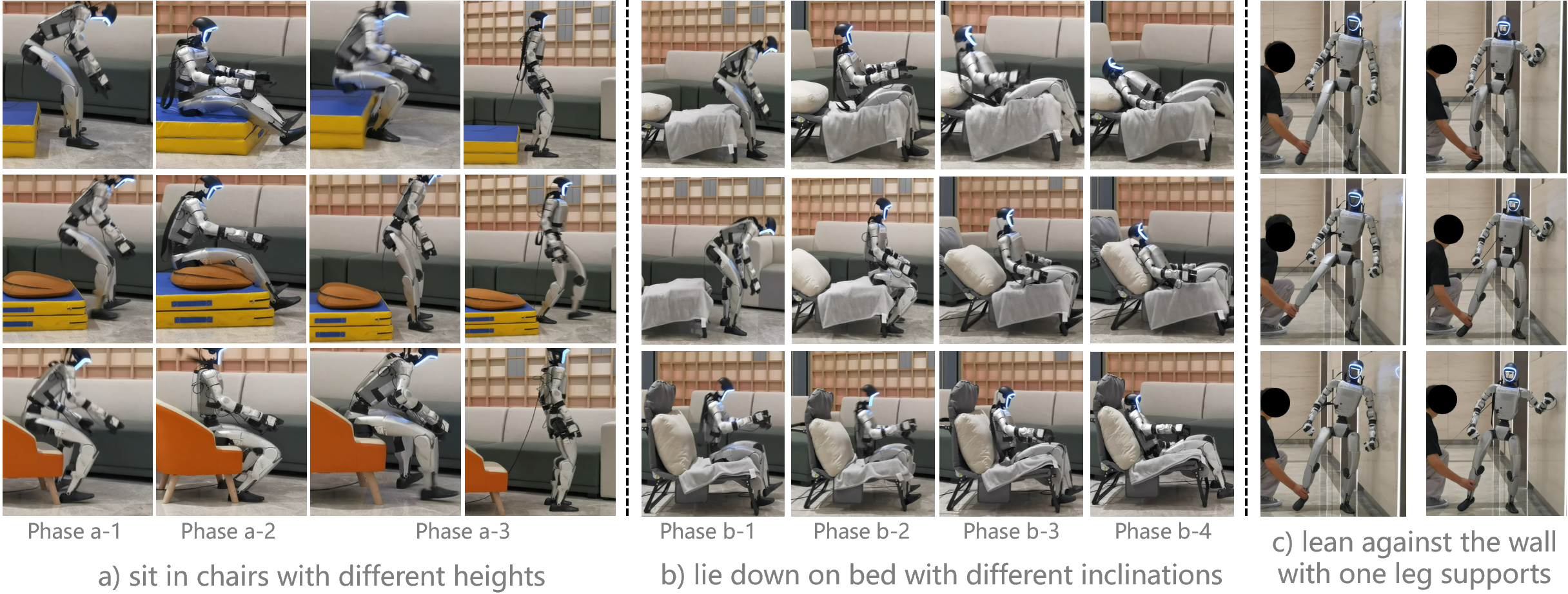}
      \captionof{figure}{
      Our work introduces a human-inspired weightlessness mechanism that controls robot joints to selectively relax when tracking non-self-stabilizing motions, enabling passive environmental contact to complete the motion. One single pipeline, without any task-specific engineering, achieves: \textbf{a) sitting on chairs} of varying heights (0.2m, 0.25m, and 0.3m); \textbf{b) lying-down on surfaces} with different inclinations (0°, 45°, and 90°); and \textbf{c) leaning against the wall} by shoulder or elbow with only one supporting leg, while the other leg resists external perturbations at varying heights (0.1m, 0.2m, and 0.3m).
      \textbf{Phase a-1, b-1}: the robot reaches a critical limit posture; \textbf{Phase a-2, b-2}: lower-limb joints relax, allowing the robot to naturally fall onto the chair or inclined surface;
      \textbf{Phase b-3}: Waist joints relax, allowing the robot's upper body to naturally fall onto the inclined surface; \textbf{Phase a-3, b-4}: Recovery to a stable, natural posture.
      }
      \label{fig: all_exp}
\end{center}
\vspace{-0.3cm}
\end{strip}

\begin{abstract}


The integration of imitation and reinforcement learning has enabled remarkable advances in humanoid whole-body control, facilitating diverse human-like behaviors. However, research on environment-dependent motions remains limited. Existing methods typically enforce rigid trajectory tracking while neglecting physical interactions with the environment. We observe that humans naturally exploit a "weightless" state during non-self-stabilizing (NSS) motions—selectively relaxing specific joints to allow passive body-environment contact, thereby stabilizing the body and completing the motion.
Inspired by this biological mechanism, we design a weightlessness-state auto-labeling strategy for dataset annotation; and we propose the Weightlessness Mechanism (WM), a method that dynamically determines which joints to relax and to what level, together enabling effective environmental interaction while executing target motions.
We evaluate our approach on 3 representative NSS tasks: sitting on chairs of varying heights, lying down on beds with different inclinations, and leaning against walls via shoulder or elbow. Extensive experiments in simulation and on the Unitree G1 robot demonstrate that our WM method, trained on single-action demonstrations without any task-specific tuning, achieves strong generalization across diverse environmental configurations while maintaining motion stability. Our work bridges the gap between precise trajectory tracking and adaptive environmental interaction, offering a biologically-inspired solution for contact-rich humanoid control.

\end{abstract}


\section{Introduction}

Developing generalist robot policies that can interact effectively with the physical world and perform complex movements has been a central goal in robotics research. 
Recent advances~\cite{fu2024humanplus,ji2024exbody2,he2025asap,he2024omnih2o,zhuang2025embrace,xie2025kungfubot} have enabled robots to perform increasingly expressive and human-like motions.

We categorize all humanoid robot motions into two major types based on whether the robot can complete the movement without assistance from the external environment or objects: the \textbf{self-stabilizing (SS) motions} and the \textbf{non-self-stabilizing (NSS) motions}. For \textbf{SS} motions, robots can complete the movements by adjusting their own postures even when interacting with the environment or objects like walk-and-turning~\cite{he2024omnih2o}, carrying boxes~\cite{ding2025humanoid}, fadeaway jumping~\cite{he2025asap}, dancing the waltz~\cite{ji2024exbody2}, and etc. For \textbf{NSS} motions, robots must rely on the external environment or objects to achieve the desired posture. Sitting on chairs, lying down on beds and leaning against walls are the most representative motions, without chairs or beds or walls, the robot struggles to maintain postural stability.

\begin{figure}[t]          
  \centering
  \includegraphics[width=1.0\columnwidth]{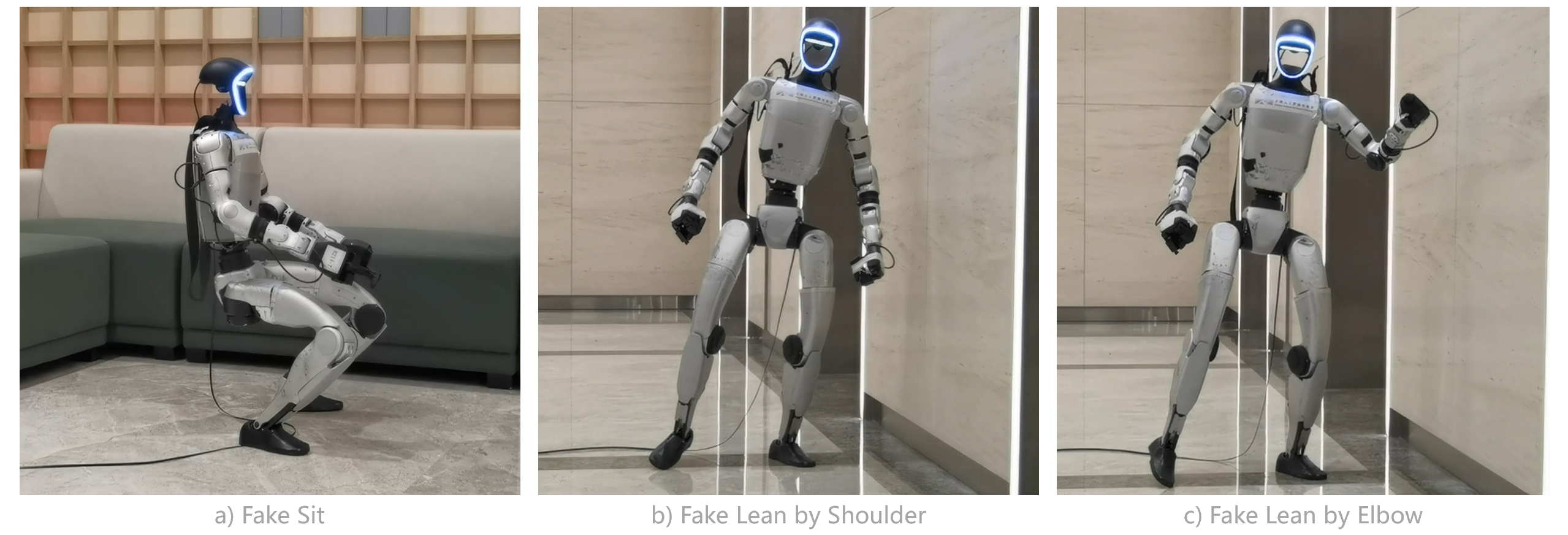}
  \caption{%
  \textbf{Fake motions.}
    Existing imitation-learning methods typically only reproduce a single demonstrated posture and ignore any physical interaction with the environment. As a result, the robot can only "fake-sit" or "fake-lean" without making sustained contact with the support surface.
  }
  \label{fig:fake}
  \vspace{-10pt}
\end{figure}

Mainstream algorithms for real robots—reinforcement learning (RL) and imitation learning (IL)—perform well on self-stabilizing motions but often fail on non-self-stabilizing ones. RL requires complex, task-specific reward design and tuning, limiting generalization to new tasks.
Naive IL produces "fake-sit" or "fake-lean" behaviors (Figure~\ref{fig:fake}), where the robot adopts visually correct postures without establishing functional contact with supporting objects. By conditioning solely on reproduced joint positions rather than physical support conditions, these policies fail to generalize when real support heights or geometries differ from training demonstrations.


\begin{figure}[h]          
  \centering
  \includegraphics[width=0.9\columnwidth]{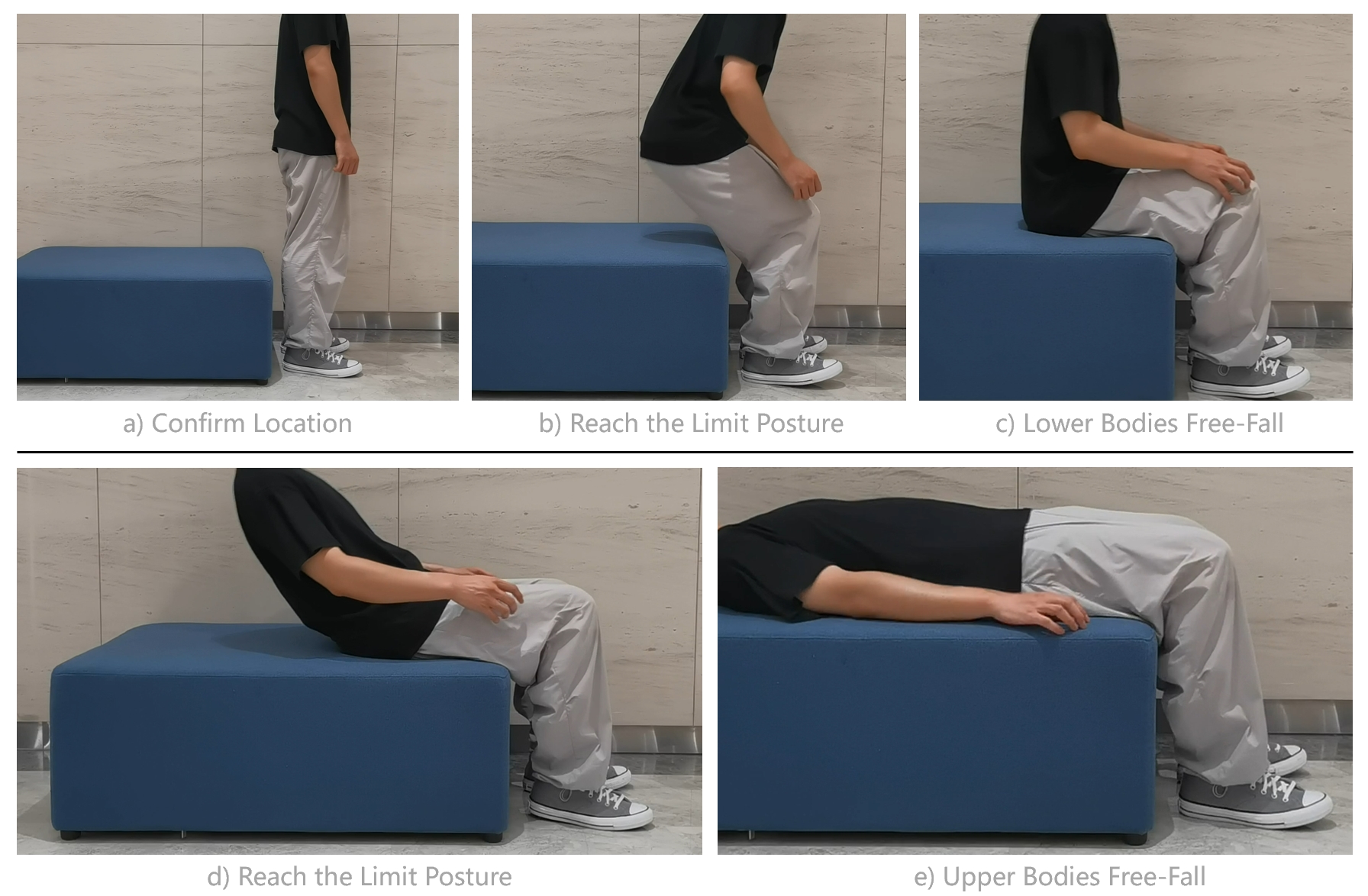}
  \caption{%
    \textbf{WM in Sitting and Lying-down.}
    During typical human sitting and lying-down motions, the body often relaxes after reaching a certain critical point. For example, after attaining a specific extreme position during sitting \textbf{b)}, the lower limbs become fully relaxed, allowing the upper body to free-fall onto the chair to complete the sitting motion. Similarly, when leaning back to a certain limit \textbf{c)}, the upper body relaxes and free-falls onto the bed to accomplish the lying-down motion.
  }
  \label{fig: p_lie}
\end{figure}

\begin{figure}[ht]          
  \centering
  \includegraphics[width=0.7\columnwidth]{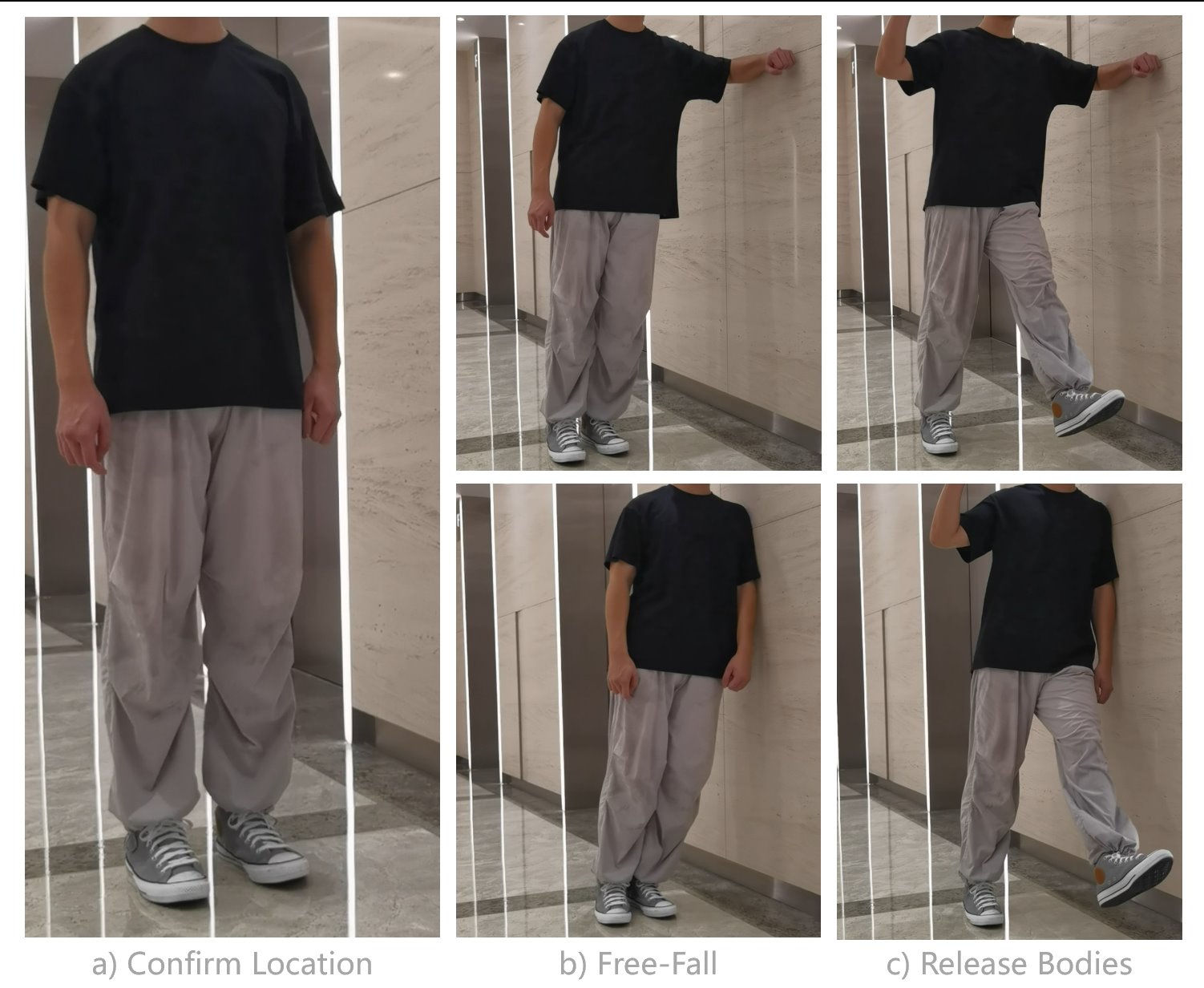}
  \caption{%
    \textbf{WM in leaning against the wall.}
    During the action of leaning against a wall, humans typically relax and undergo a controlled free-fall toward the wall after tilting to a certain critical angle in \textbf{b)}. Only the arm and leg in contact with the wall remain actively engaged to provide support, while the remaining arm and leg are free to move in \textbf{c)}.
  }
  \label{fig: p_lean}
  \vspace{-15pt}
\end{figure}

Inspired by the weightlessness mechanism in these non-self-stabilizing human motions shown in Figure \ref{fig: p_lie} and Figure \ref{fig: p_lean}, we propose a method incorporating a weightlessness mechanism (WM). This mechanism modulates joint activation by determining both which joints to relax and the specific degree of relaxation, thereby enhancing control performance during the execution of non-self-stabilizing robot movements.

To generate training data for weightless state determination, we design a weightlessness-state auto-labeling strategy to automatically determine which joints to relax and annotate their weightlessness states from motion sequences. These annotations are used to train a WM network. The trained WM network predicts the current state and appropriate joint relaxation levels, enabling the robot to execute sitting, lying-down, and leaning motions naturally through controlled free-fall. An imitation learning policy is then trained to track these motions, and fine-tuned with the WM network for more natural behaviors. The approach allows the Unitree G1 robot to sit stably on chairs of varying heights, lie down on beds with different inclinations, and lean against walls by shoulder or elbow while resisting external disturbances, as shown in Figure~\ref{fig: all_exp}.


In conclusion, our contributions are as follows:
\begin{enumerate}
[leftmargin=12pt, parsep=0pt]
\item We propose a pipeline that integrates a weightlessness mechanism (WM) into humanoid whole-body motion tracking, enabling physically plausible and human-like execution of non-self-stabilizing motions.
\item We design a weightlessness-state auto-labeling strategy that automatically annotates which joints to relax, and the weightlessness states from NSS motion sequence data.
\item We validate the approach achieves robust generalization to varied environment configurations across arbitrary NSS tasks
without any task-specific engineering.
\item We validate our approach on the Unitree G1 humanoid robot, demonstrating robust sitting on chairs of different heights, lying down onto beds with various inclinations, and leaning against walls with either the shoulder or elbow, across both simulation and real-world environments.

\end{enumerate}

\section{Related Works}

\textbf{Human-Object Interaction.}
Early datasets such as GRAB~\cite{taheri2020grab} provided detailed captures of human-object interactions. More recently, a number of datasets have been proposed to further enrich this area~\cite{bhatnagar2022behave,huang2022intercap,li2023object,kim2024parahome,liu2024mimicking,liu2024taco,lv2024himo,xie2024intertrack,liu2024core4d}.
The CHAIRS~\cite{jiang2023full} and COUCH~\cite{zhang2022couch} datasets provide realistic full-body interactions with sittable furniture, offering high-quality annotations for human–object interaction research.

Physics simulators such as Isaac Gym~\cite{makoviychuk2021isaac} and MuJoCo~\cite{todorov2012mujoco} are widely used for synthesizing realistic human-object interactions. Early methods largely rely on tracking reference motions~\cite{liu2015improving,peng2018deepmimic,chao2021learning,xie2023hierarchical,
liu2024mimicking}, but often suffer from limited generalization. To overcome this, approaches introduce Visual-Language-Action (VLA)~\cite{xu2024humanvla} and adversarial priors such as AMP~\cite{peng2021amp}, which have been applied to various tasks~\cite{juravsky2022padl,peng2022ase,hassan2023synthesizing,pan2024synthesizing,xiao2023unified}.
Recent work also explores HOI using humanoid robots to bridge simulation and reality~\cite{sferrazza2024humanoidbench,liu2024mimicking,allshire2025visual}.


\textbf{Imitation Learning on Humanoid.}
Recent advances in humanoid control have shifted from model-based methods to perception- and imitation-driven approaches. Early work relied on IK, ZMP, and LIP models for teleoperation~\cite{montecillo2010real,hu2014online},
later extended by motion capture~\cite{stanton2012teleoperation,dajles2018teleoperation}, Exoskeleton~\cite{ben2025homie}, and VR-based systems~\cite{winkler2022questsim,lee2023questenvsim,ye2022neural3points,luo2023universal,cheng2024open}. 

Deep learning enables large-scale motion imitation. Humanoid Transformer~\cite{radosavovic2024humanoid} predicts motions from mocap and video data, and Exbody~\cite{cheng2024expressive} enhances expressiveness via body-part separation. HumanPlus~\cite{fu2024humanplus} and Okami~\cite{li2024okami} extend video imitation with transformer models. BeyondMimic~\cite{truong2025beyondmimic} achieves real-world dynamic skills with guided diffusion, and GBC~\cite{yao2025gbc} introduces a robust DAgger-MMPPO based MMTransformer. Real-time control methods such as Exbody2~\cite{ji2024exbody2}, H2O~\cite{he2024learning}, OmniH2O~\cite{he2024omnih2o}, and HOVER~\cite{he2024hover} use teacher-student frameworks for teleoperation and imitation. To address motion diversity loss, UniTracker~\cite{yin2025unitracker} integrates a CVAE to model latent motion diversity. For expressive motions, Humanoid-VLA~\cite{ding2025humanoid} adds scene perception for whole-body control, while ASAP~\cite{he2025asap}, Embrace Collisions~\cite{zhuang2025embrace}, and PBHC~\cite{xie2025kungfubot} enhance expressiveness via delta-action pretraining, contact recovery, and dynamic skill imitation.
\section{Methods}
\begin{figure*}[ht]
  \centering
  \includegraphics[width=\linewidth]{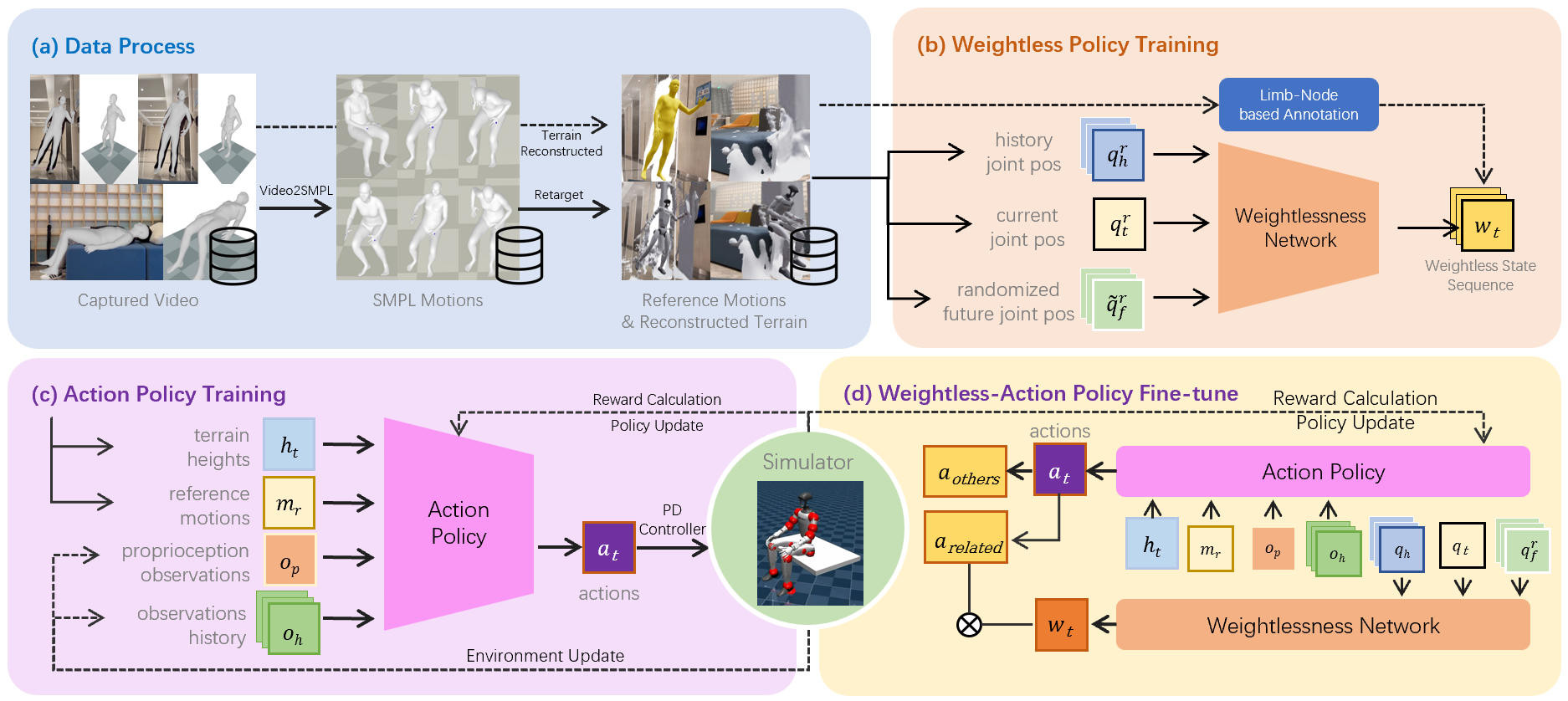}
  \caption{\textbf{Framework.} The method consist of 4 components:
  \textbf{(a)} We collected videos of sitting, lying, and leaning (shoulder and elbow) motions, from which SMPL motions and terrains were extracted, followed by motion retargeting and terrain reconstructed.
\textbf{(b)} Label the robot's joint weightlessness states and weightless time interval via annotation method, and train a WM network using LSTM.
\textbf{(c)} Train the action policy using imitation learning.
\textbf{(d)} Integrate the WM network to fine-tune the learned policy.}
  \label{fig:framework}
  \vspace{-15pt}
\end{figure*}
\subsection{Framework}

The proposed framework, incorporating a human-inspired weightlessness mechanism (WM), is shown in Figure~\ref{fig:framework}. We collected 50 video demonstrations each of sitting, lying down, and leaning with shoulder/elbow motions, retargeted them to the Unitree G1 robot. Environmental supports (e.g., chairs, beds, walls) were reconstructed into height map, and joint weightlessness states were annotated across the dataset.

We adopt a two-stage training for non–self-stabilizing motions. First, an asymmetric Proximal Policy Optimization (PPO) algorithm performs imitation learning on the processed data. Second, an LSTM-based WM network predicts continuous joint relax levels, simulating controlled free-fall. The imitation policy is fine-tuned under these signals to improve naturalness and robustness.

The final policy was validated in MuJoCo and deployed on the Unitree G1, successfully executing non–self-stabilizing motions. Training details are provided in later sections.

\subsection{Data Process}

\textbf{Data Acquisition and Processing.} We collected 200 video demonstrations across four action categories—sitting, lying, and leaning via shoulder and elbow (50 per category). SMPL motion data were obtained using GVHMR~\cite{shen2024worldgrounded}, world point clouds extracted with MegaSaM~\cite{li2024megasam}, and meshes generated via NKSR~\cite{huang2023nksr}. SMPL motions were aligned with the reconstructed scene geometry using the MegaHunter module from VideoMimic~\cite{allshire2025visual}, after which we approximated the key contact objects and environment using simple shape of boxes.

\textbf{Retargeting with Smoothing.} We adopt GMR \cite{gmr2025} retargeting~\cite{luo2023perpetual} to transfer reconstructed SMPL motions onto target skeletons of Unitree G1. To alleviate jitter in static poses, we apply a smoothing pipeline that first downsamples the sequence to reduce redundancy, then applies a causal moving average to suppress high-frequency oscillations, and finally uses a median filter to remove spike-like artifacts while preserving salient motion features. Interpolation restores the original frame rate, producing temporally consistent and stable motions for downstream optimization.

\subsection{Weightlessness-State Auto-Labeling Strategy}
To accurately annotate weightless states in motion sequences, we identify two key aspects: (i) the time interval of weightlessness, and (ii) the set of joints entering the weightless state.  

\begin{figure}[ht]          
  \centering
  \includegraphics[width=0.9\columnwidth]{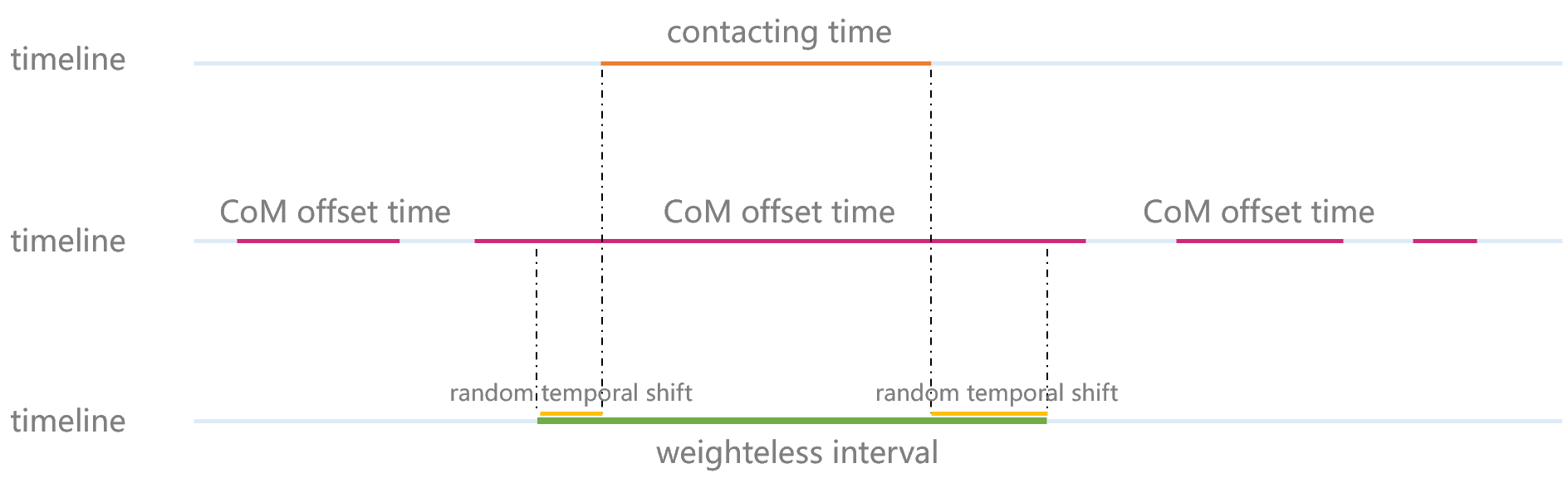}
  \caption{
  \textbf{Confirmation of Weightless Time Interval.}
  The weightless interval $\mathcal{I}$ is related to contacting-time $\mathcal{C}(t)$, CoM offset time $\{\mathcal{P}(t)\notin\mathcal{S}(t)\}$, and random temporal shift $\Delta t$.
  }
  \label{fig: time interval}
  \vspace{-15pt}
\end{figure}

\textbf{Time Interval of Weightlessness.}
Let $\mathcal{P}(t)$ denote the gravity projection of the robot’s center of mass at time $t$, and $\mathcal{S}(t)$ the support polygon formed by the feet. A weightless interval should be satisfied whenever
\[
\mathcal{P}(t) \notin \mathcal{S}(t),
\]
and the robot makes external contact through body parts other than the feet. Let $\mathcal{C}(t)$ denote the set of environmental contacts at time $t$, which is confirmed in reconstructed simulator. Then the annotated weightless interval $\mathcal{I}$ is 
\[
\mathcal{I} = \{t ~|~ \mathcal{P}(t) \notin \mathcal{S}(t), ~\mathcal{C}(t) \neq \emptyset \}.
\]
To enhance robustness, the start and end times of $\mathcal{I}$ are perturbed by a random temporal shift of $\pm \Delta t$, with $\Delta t = 20$ timesteps. The interval is also defined as shown in Figure~\ref{fig: time interval}.

\textbf{Joint Weightless States.}
The set of active joints $\mathcal{A}(t)$ is determined using a limb-node hierarchy inspired by human motor control. Let $\mathcal{C}(t)$ again denote the set of environment-contact frames at time $t$. Then
\[
\mathcal{A}(t) = \{c_\text{waist}\} \cup \bigcup_{c \in \mathcal{C}(t)} \mathrm{ParentPath}(c),
\]
where $\mathrm{ParentPath}(c)$ denotes all ancestor nodes from contact joint $c$ up to the waist joint $c_\text{waist}$. The weightless joints are then
\[
\mathcal{W}(t) = \{1,2,\dots,K\} \setminus \mathcal{A}(t).
\]

\begin{figure}[ht]          
  \centering
  \includegraphics[width=0.9\columnwidth]{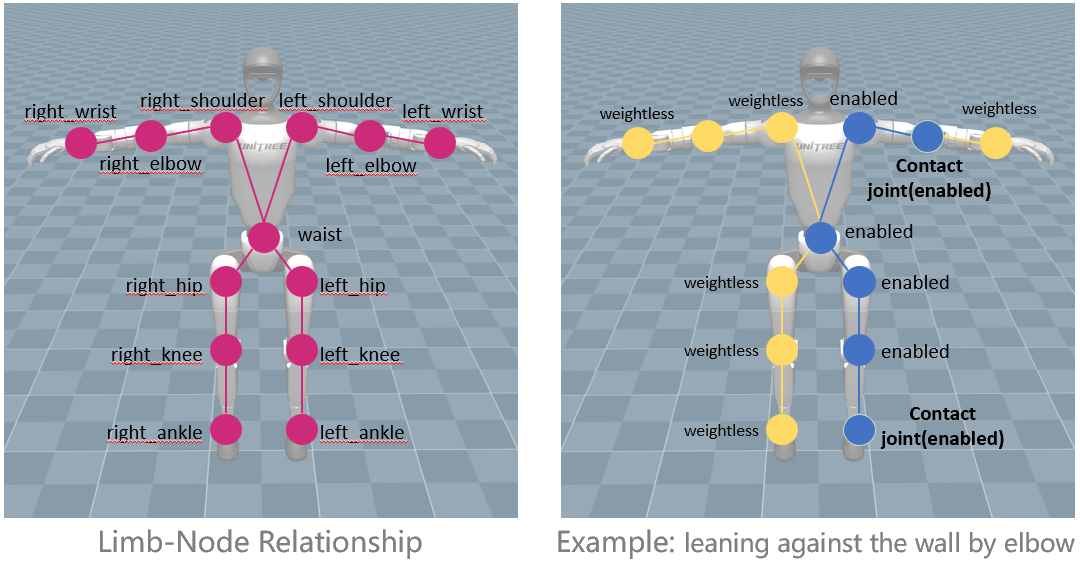}
  \caption{
  \textbf{Confirmation of Joint Weightless States.}
  The left image illustrates the limb-node relationship. The right image presents a case study.
  }
  \label{fig: joint_idx}
  \vspace{-15pt}
\end{figure}

The method and example were shown in Figure~\ref{fig: joint_idx}, when leaning against a wall supported by the left elbow and left foot, we have $\mathcal{C}(t)=\{c_\text{left elbow}, c_\text{left foot}\}$, hence $\mathcal{A}(t)$ includes all ancestors of these contacts plus the waist (painted blue), while the remaining joints $\mathcal{W}(t)$ are annotated as weightless (painted red).

Through the aforementioned approach, we can achieve automated annotation of weightless time interval and weightless joints directly from video stream data, eliminating the need for manual intervention.

\subsection{WM Network Training} 

The weightlessness mechanism (WM) computes joint-specific relaxation levels $w_{t,i}$ for the corresponding joints $q_i$, enabling the robot to transition into a weightless state and execute non-self-stabilizing tasks.

Formally, let the robot have $K=23$ actuated joints with joint states $q = [q_1, q_2, \dots, q_K]^\top \in \mathbb{R}^K$, where $q_i$ represents the angle or position of the $i$-th joint. The WM aims to learn a continuous control policy
\[
W_\theta: \mathcal{X} \to [0,1]^K,
\]
that maps observed features $\mathbf{x}_t \in \mathcal{X}$, consisting of current posture, motion history, and future motion trend, to a target relaxation vector
\[
\mathbf{w}_t = [w_{t,1}, w_{t,2}, \dots, w_{t,K}]^\top,
\]
where $w_{t,i} \in [0,1]$ denotes the relaxation level of joint $i$ at time $t$. A value $w_{t,i} \approx 0$ indicates full relaxation (weightless), while $w_{t,i} \approx 1$ indicates full activation.

To determine the weightlessness state, it is necessary to utilize the current posture, motion history, and future motion trend. In particular, future motion information is crucial.

Let $q^r_t \in \mathbb{R}^{23}$ denote the reference joint positions at time $t$, and let
\[
q^r_h(t) = \{q^r_{t-h+1}, \dots, q^r_{t-1}\}
\]
denote the historical sequence of length $h=4$. The future motion trend is obtained by sampling a reference sequence with a randomized temporal offset. Specifically, for maximum offset $\tilde{t}$ and horizon $l=5$, we define
\[
\tilde{q}^r_f(t) = \{q^r_{t+\delta},~ q^r_{t+\delta+1},~ \dots,~ q^r_{t+\delta+l-1}\}
\]
where $\delta \sim \mathcal{U}(-\tilde{t},~\tilde{t})$ is uniformly sampled.Thus, the network input at time $t$ is
\[
\mathbf{x}_t = \big[q^r_h(t),~ q^r_t,~ \tilde{q}^r_f(t)\big] \in \mathbb{R}^{230},
\]
corresponding to $4+1+5=10$ frames of $K$-dimensional joint positions.  

The LSTM-based weightlessness mechanism network $W_\theta$ maps this input to a continuous relaxation vector:
\[
\mathbf{w}_t = W_\theta(\mathbf{x}_t) \in [0,1]^K,
\]

To enforce temporal consistency and avoid abrupt changes in relaxation levels, we add a smoothness regularization term across consecutive frames. Specifically, for each training sequence, we define
\[
\mathcal{L}_{\text{smooth}}(\theta) 
= \frac{1}{K}\sum_{i=1}^K \sum_{t} \big( w_{t,i} - w_{t-1,i} \big)^2,
\]
which penalizes sudden jumps in predicted relaxation levels between adjacent frames.  

The final training objective is a weighted sum of the Binary Cross-Entropy (BCE) loss and the smoothness loss:
\[
\mathcal{L}_{\text{total}}(\theta) 
= \mathcal{L}_{\text{BCE}}(\theta) + \lambda \, \mathcal{L}_{\text{smooth}}(\theta),
\]
where $\lambda > 0$ controls the strength of temporal smoothing. This ensures that the predicted relaxation vector $\mathbf{w}_t$ evolves continuously over time, preventing unrealistic sudden transitions of relaxation level.

Details of the network architecture and training configurations are included in Table~\ref{tab:lstm_config}.

\begin{table}[htbp]
\centering
\caption{LSTM Network Configurations}
\label{tab:lstm_config}
\resizebox{\linewidth}{!}{%
    \begin{tabular}{ll}
    \toprule
    \textbf{Component} & \textbf{Specification} \\
    \midrule
    Input Dimension & 230-dims each frame \\
     & including 4 historical steps, 1 current step,\\
     & and 5 future steps of joint positions $q\in\mathbb{R}^{23}$,\\
     & resulting in a total input dimension of $23\times10=230$ \\
    Hidden Size & [256, 256, 64] units for hidden layers \\
    Output & Continuous deactivation levels $\mathbf{w}_t \in [0,1]^K$ \\
    Activation Function & Sigmoid \\
    Dropout & 0.1 \\
    Optimizer & Adam \\
    Loss Function & Binary Cross-Entropy (BCE) + Smoothness penalty \\
    Batch Size & 64 \\
    Learning Rate & 1e-4 \\
    Regularization & Temporal smoothness: $\lambda \sum (w_{t,i} - w_{t-1,i})^2$ \\
    \bottomrule
    \end{tabular}
}
    \vspace{-10pt}
\end{table}

\subsection{Action Policy Training}
In this work, imitation learning is primarily used to train an action policy for the robot's whole-body motion. We employ the policy network that maps from the robot's proprioception and reference motions to whole-body motion control signals.

 \textbf{Observation space.} The policy observation $o_t$ consists of joint positions $q_t \in \mathbb{R}^{23}$, joint velocities $\dot{q}_t \in \mathbb{R}^{23}$, root angular velocity $\omega^{\text{root}}_t \in \mathbb{R}^3$, projected gravity $g_t \in \mathbb{R}^3$ and previous action $a_{t-1} \in \mathbb{R}^{23}$. To enhance the tracking performance, we incorporate the reference joint positions \( q^{r}_t \in \mathbb{R}^{23} \) and joint velocities \( \dot{q}^{r}_t \in \mathbb{R}^{23} \) extracted from the dataset into the observations. Additionally, we introduce an n - step observation history $o_{t-n : t-1}$ for more precise future state estimation. To improve terrain adaptation, we further include sampled terrain height measurements \( h_t \in \mathbb{R}^{32} \) around the robot's base. We sample 32 points within a region of 0.5×0.5 meters from the robot's base position, rotated according to the robot's orientation.

Thus, the complete observation at time $t$ is:
\[
\mathbf{o}_t = [\ q_{t} \quad \dot{q}_{t} \quad \omega^{\text{root}}_{t} \quad g_{t} \quad a_{t-1} \quad q^{r}_t \quad \dot{q}^{r}_t  \quad h_t \quad o_{t-n : t-1} \ ]
\]


We adopt an asymmetric PPO framework, where the actor and critic use distinct observation spaces. The actor receives only partial observations $\mathbf{o}_t$ available at deployment, while the critic has access to additional privileged information, including root linear velocity $v^{\text{root}}_t \in \mathbb{R}^3$, ground friction $\mu \in \mathbb{R}^1$, and external force $f \in \mathbb{R}^3$. This enables more accurate value estimation, improving overall policy optimization.

\textbf{Action Space.} The policy outputs an action $a_t \in \mathbb{R}^{K}$, corresponding to the Unitree G1's 23 degrees of freedom. Each element of $a_t$ represents the deviation between the desired joint position and the default joint position. The desired joint positions are subsequently tracked using a low-level Proportional-Derivative (PD) controller, which generates the torque commands to actuate.

\begin{table}[htbp]
    \centering
    \caption{Reward Terms for Training}
    \label{table:reward}
    \resizebox{\linewidth}{!}{%
    \begin{tabular}{@{}llll@{}}
        \toprule
        & \textbf{Term} & \textbf{Expression} & \textbf{Weight} \\
        \midrule
        
        \multicolumn{4}{@{}l}{\textbf{Task Rewards}} \\
        \midrule[0.5pt]
        & Keypoint position & \( \exp(-0.1 \|p_t - \hat{p}_t\|_2^2) \) & $3.00$ \\
        & Root rotation & \( \exp(-1.0 \|\theta^{root}_t \oplus \hat{\theta}^{root}_t\|) \) & $0.5$ \\
        & Root velocity & \( \exp(-1.0 \|v^{root}_t - \hat{v}^{root}_t\|_2) \) & $0.75$ \\
        & Joint position without feet & \( \exp(-1.0 \|d_t - \hat{d}_t\|_2^2) \) & $32$ \\
        & Joint velocity without feet & \( \exp(-1.0 \|\dot{d}_t - \hat{\dot{d}}_t\|_2^2) \) & $0.5$ \\
        & Termination & \( \mathbf{1}_{\text{termination}} \) & $-200$ \\
        \midrule[0.5pt]
        
        \multicolumn{4}{@{}l}{\textbf{Regularization Rewards}} \\
        \midrule[0.5pt]
        & Joint acceleration & \( \|\ddot{q}_t\|_2 \) & $-2.5\times10^{-8}$ \\
        & Joint velocity & \( \|\dot{q}_t\|_2 \) & $-0.001$ \\
        & Action rate & \( \|a_t - a_{t-1}\|_2 \) & $-0.5$ \\
        & Torque & \( \|\tau_t\| \) & $-1\times10^{-6}$ \\
        & Feet orientation & \( \left\|\mathbf{g}_{xy}\right\|_2 \) & $-62.5$ \\
        & Feet heading & \( \|\theta^{feet}_t - \theta^{root}_t\|_2 \) & $-1\times10^{-5}$ \\
        \bottomrule
    \end{tabular}
    }
    \vspace{-10pt}
\end{table}

\renewcommand{\thetable}{\Roman{table}}

\textbf{Reward Design.}
During the first stage, we employed a reward structure inspired by \cite{he2025asap}. However, for non-self-stabilizing motions, we observed noticeable inconsistencies in the feet' placement after retargeting. To address this issue, we opted to relax the tracking constraints on the feet joints during training and introduced additional rewards for feet orientation and feet heading. This approach enhances the stability and improves its tracking performance for key points. The detailed reward design is summarized in Table~\ref{table:reward}.

\subsection{Weightless-Action Policy Fine-tune.}
At the second-stage training, we integrate the weightlessness mechanism to fine-tune the action policy.
The WM network predicts the degree of deactivation for each target lower-limb joint at each step. 
The output torque of the robot's joints can be expressed as:
\[
    \tau_t = \tau(a_{t})\cdot \mathbf{w}_t
\]
where is, \( \tau_t\in\mathbb{R}^{23} \) represents the joint output torque at the current moment, \(\tau(\cdot) \) represents the torque mapping of the PD controller. $a_{t}\in\mathbb{R}^{23}$ represents the action policy output for the whole-body joints.

\begin{table}[ht]
    \centering
    \caption{Ablation Study: Weightlessness Mechanism Effectiveness (Tracking Metrics)}
    \small
    \label{table:ablation_weightless_error}
    \resizebox{0.9\linewidth}{!}{%
    \begin{tabular}{c|p{0.8cm}p{0.8cm}p{0.8cm}p{0.8cm}p{0.8cm}p{0.8cm}}
        \toprule
        \multirow{2}{*}{Experiments}& \multicolumn{6}{c}{Error Metrics}\\
        \cmidrule(lr){2-7}
             & $E_{mpjpe}$ & $E_{mpjae}$ & $E_{mpjve}$ & $E_{root,p}$ & $E_{root,r}$ & $E_{root,v}$ \\
        \midrule
        baseline-Sit & \textbf{0.1642} & \textbf{0.0357} & 1.7802 & \textbf{0.1468} & \textbf{1.4936} & 0.3350 \\
        WM w/o ft            & 0.2047 & 0.0421 & 1.9171 & 0.1734 & 1.2011 & 0.5302 \\
        WM+ft (ours)     & 0.1707 & 0.0398 & \textbf{1.7484} & 0.1521 & 1.4550 & \textbf{0.3099} \\
        \midrule
        baseline-LieDown & \textbf{0.2512} & \textbf{0.0483} & 2.1504 & \textbf{0.2317} & 1.8901 & \textbf{0.4103} \\
        WM w/o ft & 0.2953 & 0.0556 & 2.4505 & 0.2651 & \textbf{1.7523} & 0.6226 \\
        WM+ft (ours) & 0.2605 & 0.0505 & \textbf{2.1054} & 0.2380 & 1.8329 & 0.4357 \\
        \midrule
        baseline-Lean Shoulder & 0.1378 & \textbf{0.0281} & 1.2546 & \textbf{0.0957} & 0.8758 & 0.1832 \\
        WM w/o ft & 0.1158 & 0.0332 & 1.4891 & 0.1176 & 0.7926 & 0.2783 \\
        WM+ft (ours) & \textbf{0.1103} & 0.0291 & \textbf{1.2054} & 0.0987 & \textbf{0.8552} & \textbf{0.1959} \\
        \midrule
        baseline-Lean Elbow  & \textbf{0.1084} & \textbf{0.0275} & 1.2203 & \textbf{0.0927} & \textbf{0.8487} & 0.1884 \\
        WM w/o ft  & 0.1328 & 0.0327 & 1.4592 & 0.1114 & 0.7339 & 0.2920 \\
        WM+ft (ours) & 0.1127 & 0.0283 & \textbf{1.1809} & 0.0949 & 0.8554 & \textbf{0.1839} \\
        \bottomrule
    \end{tabular}
    }
\end{table}

\begin{table}[ht]
    \centering
    \caption{Ablation Study: Weightlessness Mechanism Effectiveness (Success Rates)}
    \small
    \label{table:ablation_weightless_success}
    \resizebox{0.7\linewidth}{!}{%
    \begin{tabular}{c|p{0.8cm}p{0.8cm}p{0.8cm}}
        \toprule
        \multirow{2}{*}{Experiments}& \multicolumn{3}{c}{Success Rates}\\
        \cmidrule(lr){2-4}
             & $E_{succ}^{1}$ & $E_{succ}^{2}$ & $E_{succ}^{3}$\\
        \midrule
        baseline-Sit & 0.72 & 0.20 & 0.08 \\
        WM w/o ft          & 0.76 & 0.86 & 0.82 \\
        WM+ft (ours)     & 0.78 & 0.92 & 0.84\\
        \midrule
        baseline-LieDown & 0.58 & 0.56 & 0.64 \\
        WM w/o ft & 0.72 & 0.72 & 0.74 \\
        WM+ft (ours) & \textbf{0.92} & \textbf{0.88} & \textbf{0.90} \\
        \midrule
        baseline-Lean Shoulder & 0.48 & 0.30 & 0.22 \\
        WM w/o ft & 0.70 & 0.62 & 0.62 \\
        WM+ft (ours) & \textbf{0.94} & \textbf{0.88} & \textbf{0.82} \\
        \midrule
        baseline-Lean Elbow  & 0.42 & 0.24 & 0.14 \\
        WM w/o ft  & 0.86 & 0.84 & 0.68 \\
        WM+ft (ours) & \textbf{0.92} & \textbf{0.92} & \textbf{0.86} \\
        \bottomrule
    \end{tabular}
    }
    \vspace{-8pt}
\end{table}

The input to the WM network while fine-tuning can be expressed as:
\begin{equation}
    \mathbf{w}_t = W_\theta[q_h,~q_t,q^r_f]
\end{equation}
where $q_t$ and $q_h$ represent the current and historical joint positions of the robot, respectively, while $q^r_f$ corresponds to the future reference motion initialized from the current time phase. Meanwhile, the incorporation of the weightlessness mechanism alters the observation space and may impair the performance of the original policy, thus making fine-tuning of the policy essential.

\section{Experiments}
\subsection{Experiment Setup}
\textbf{Software and Hardware.} To verify the effectiveness of the algorithm, we deployed it in both simulation and real-world environments. We used Isaac Gym as the training environment and validated our policy in the MuJoCo simulator. Both the simulation and real-world experiments were conducted using the Unitree G1 robot. We utilize the built-in LiDAR on the Unitree G1 robot to obtain terrain height observations. The control frequency on the real robot is 50 Hz.

\textbf{Baselines.} We use the imitation learning policy without the WM as our baseline and perform ablation experiments on this baseline to evaluate the effectiveness of the proposed mechanisms.

\textbf{Metrics.} We evaluate our method using a set of metrics that assess kinematic accuracy, physical realism, and task success in simulation. $E_{\text{mpjpe}}$, $E_{\text{mpjae}}$, $E_{\text{mpjve}}$: These metrics evaluate the accuracy of joint-level predictions. $E_{\text{mpjpe}}$ measures the mean per-joint global position error, $E_{\text{mpjae}}$ measures the mean per-joint angles error, and $E_{\text{mpjve}}$ measures the mean per-joint velocity error. \text{$E_{\text{root,p}}$, $E_{\text{root,r}}$, $E_{\text{root,v}}$}: These metrics quantify the root tracking performance in terms of position, rotation, and velocity error, respectively. $E_{\text{succ}}^{1}$, $E_{\text{succ}}^{2}$, and $E_{\text{succ}}^{3}$ denote the task success rates measured in four different experimental motions:

\begin{itemize}
    \item \textbf{Sit action:} The success rates $E_{\text{succ}}^{1}$, $E_{\text{succ}}^{2}$, and $E_{\text{succ}}^{3}$ correspond to the agent successfully sitting on chairs of three different heights: 0.2\,m, 0.25\,m, and 0.3\,m. A trial is considered successful if the agent achieves the sitting posture and contact with chairs without falling.
    
    \item \textbf{Lying down action:} The success rates $E_{\text{succ}}^{1}$, $E_{\text{succ}}^{2}$, and $E_{\text{succ}}^{3}$ for the lying-down action are measured at three tilt angles: $0^\circ$, $45^\circ$, and $90^\circ$. A trial is considered successful if the agent completes the lying-down motion as intended, maintaining stability without any abnormal oscillations or jittering.

    \item \textbf{Leaning action (by shoulder/elbow)}:
    The leaning task is designed to test whether the agent can maintain stability while one side of the body bears weight against a support via the shoulder or elbow. The success rates $E_{\text{succ}}^{1}$, $E_{\text{succ}}^{2}$, and $E_{\text{succ}}^{3}$ are defined by the free swing and disturbance resistance of the non-weight-bearing leg, tested at raised heights of 0.1m, 0.2m, and 0.3m.
\end{itemize}

\noindent
For each condition, the reported success rates are computed based on 50 trials.

\subsection{Main Results}
In this work, our main contribution is the introduction of a WM. To evaluate its effectiveness, we focus on these core questions:



\textbf{Q1: Does the WM benefit?}

To evaluate the benefits of the WM in training, we introduce it as an experimental variable. As shown in Table~\ref{table:ablation_weightless_error}, the inclusion of WM leads to a slight degradation in motion tracking performance, as expected, since it permits controlled deviation from the reference trajectory to achieve more natural motion. However, the baseline method, which learns a fixed motion trajectory, fails to adapt to environmental variations such as chair height or surface inclination. In contrast, as shown in Table~\ref{table:ablation_weightless_success}, the WM-based approach significantly improves success rates across all tasks. These results indicate that although WM slightly affects tracking accuracy, it substantially enhances the robot's ability to complete sitting, lying, and leaning motions successfully.

\begin{figure}[t]          
  \centering
  \includegraphics[width=1.0\columnwidth]{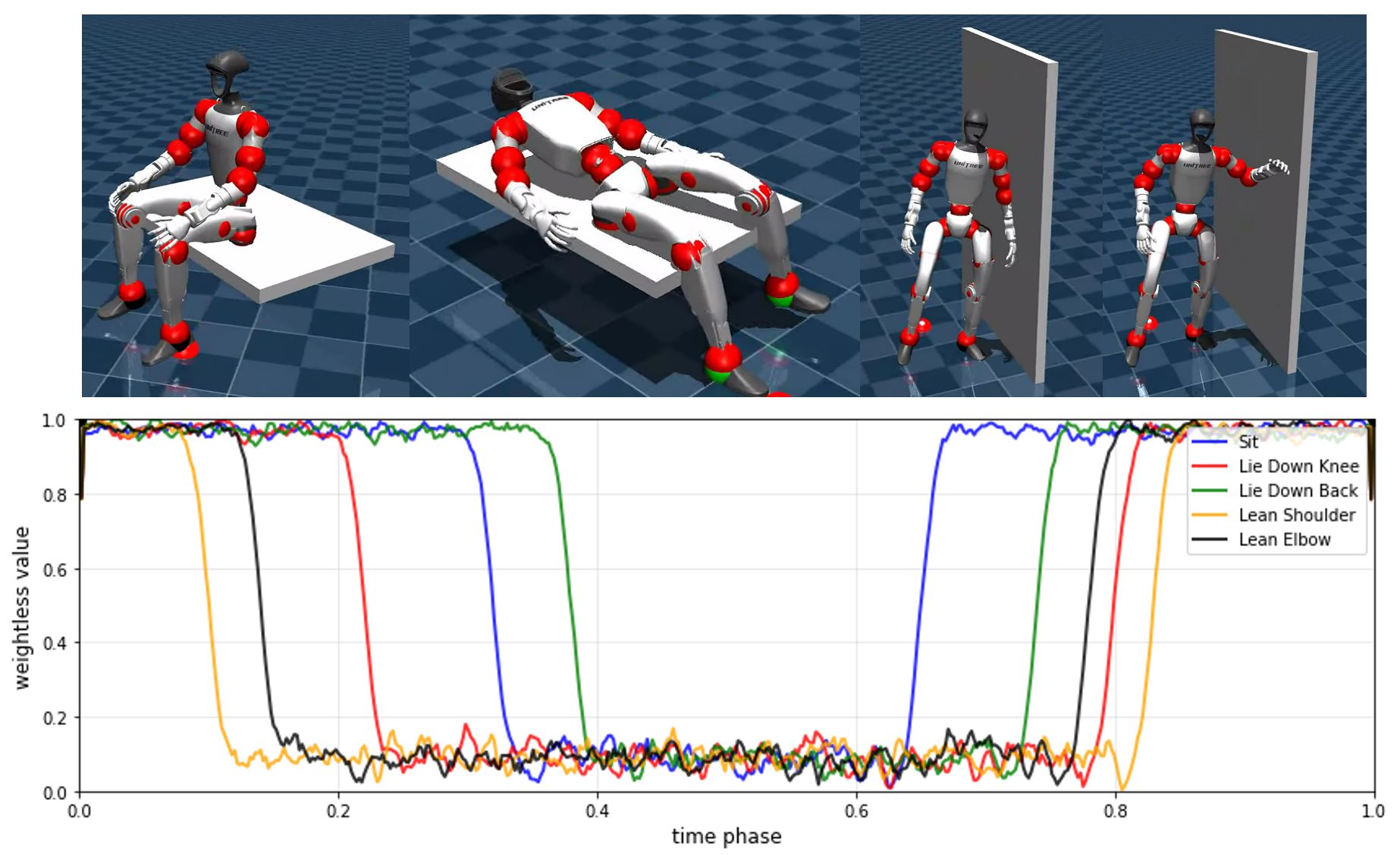}
  \caption{%
    For different motions, the weightlessness network demonstrates a clear capability to identify distinct temporal intervals of weightlessness.
  }
  \label{fig: output}
  \vspace{-10pt}
\end{figure}

\textbf{Q2: Why is the WM used only for fine-tuning rather than during initial policy training?}

Integrating the independently trained WM into imitation learning poses a challenge. We designed an ablation experiment where the way WM involves in training differ: applying WM from the start (WM w/o ft) or applying WM to fine-tune the action policy after initial convergence (WM+ft).

As results shown in Table~\ref{table:ablation_weightless_error} and Table~\ref{table:ablation_weightless_success}, fine-tuning 5000 iterations of imitation learning followed by 2000 iterations with the WM (WM+ft) achieves better tracking and success rates than synchronous training (WM w/o ft). This is likely because early random exploration triggers premature weightless states, leading to overly conservative behavior. Fine-tuning avoids this, allowing better overall performance.

\begin{table}[htbp]
\centering
\caption{Domain Randomization Parameters}
\label{tab:domain_randomization}
\begin{tabular}{lcl}
\toprule
\textbf{Parameter} & \textbf{Range [Min, Max]} & \textbf{Unit} \\
\midrule
Link Mass & $[0.8, 1.2]$  & kg \\
CoM offset & $[-0.1, 0.1]$ & m \\
Ground Friction & $[0.5, 1.5]$ & - \\
Motor Strength & $[0.8, 1.2]$ & Nm \\
Joint $K_p$ & $[0.75, 1.25]$ & - \\
Joint $K_d$ & $[0.75, 1.25]$ & - \\
Action Delay & $[5, 25]$ & ms\\
\bottomrule
\end{tabular}
\vspace{-10pt}
\end{table}

\textbf{Q3: How does WM network perform?}

To evaluate the performance intuitively, we output the weightless values for four different motions, as shown in the Figure \ref{fig: output}. It can be observed that during the motion, the weightlessness network successfully identifies reasonable temporal intervals of weightlessness.

\subsection{Deployment}
To bridge the gap between simulation and reality (sim-to-real), domain randomization was introduced during training. By introducing variability in key physical parameters, the learned policy becomes more robust, enabling seamless transfer to the real world. The specific parameters used for domain randomization are listed in Table~\ref{tab:domain_randomization}.

As shown in Figure~\ref{fig: all_exp}, the proposed algorithm was deployed on the real robot.
For sitting, the robot was tested under different chair heights and achieved the intended posture reliably.
For lying-down, trials under varying surface inclinations were all completed stably without abnormal oscillations.
For leaning, both shoulder- and elbow-based cases were evaluated against a wall, with stability confirmed by perturbing the non-supporting leg.
Overall, the robot demonstrated consistent, natural, and stable motion performance across all conditions.

\section{Conclusion}


In this work, we introduce a human-inspired weightlessness mechanism (WM) designed to enable humanoid robots to perform physically plausible non-self-stabilizing (NSS) motions—including sitting on chairs of varying heights, lying down on bed with different inclinations, and leaning against walls by shoulder or elbow. A auto-labeling strategy based on limb-node relationship is proposed, which enables automatic annotation of joint weightless states and weightless time interval. The proposed WM network employs an LSTM-based network to predict adaptive, joint-specific relaxation levels in real time, allowing the robot to enter a state of controlled free-fall while preserving body stability. Extensive experiments conducted on the Unitree G1 humanoid robot show significantly improved task success rates and more natural NSS motion adaptation across diverse environmental geometries without any task-specific engineering. This approach enhances the robot's ability to imitate human-like weight transfer and whole-body control in interactive scenarios.

\bibliography{root}

\end{document}